\documentclass[letterpaper]{article} 
\usepackage{aaai25}  
\usepackage{times}  
\usepackage{helvet}  
\usepackage{courier}  
\usepackage[hyphens]{url}  
\usepackage{graphicx} 
\usepackage{natbib}  
\usepackage{caption} 
\usepackage{algorithm}
\usepackage{algorithmic}

\usepackage{newfloat}
\usepackage{listings}

\usepackage{dsfont}
\usepackage{xcolor}
\usepackage{amsmath}
\usepackage{amssymb}
\usepackage{enumitem}

\DeclareCaptionStyle{ruled}{labelfont=normalfont,labelsep=colon,strut=off} 
\floatstyle{ruled}
\newfloat{listing}{tb}{lst}{}
\floatname{listing}{Listing}
\newcommand{\argmax}{\mathop{\mathrm{argmax}}\limits}

\usepackage[capitalise,nameinlink]{cleveref}
\usepackage{multirow}

\newlist{myenum}{enumerate}{1}
\setlist[myenum]{label=(\arabic*)}

\crefname{myenumi}{}{}
\Crefname{myenumi}{}{}

\title{TransferLight: Zero-Shot Traffic Signal Control on any Road-Network}
\author{
    Johann Schmidt\equalcontrib\textsuperscript{\rm 1},
    Frank Dreyer\equalcontrib,
    Sayed Abid Hashimi,
    Sebastian Stober
}
\affiliations{
    Artificial Intelligence Lab \\
    Otto-von-Guericke University \\
    Magdeburg, Germany \\
    \textsuperscript{\rm 1} johann.schmidt@ovgu.de
}

\begin{document}

\maketitle

\begin{abstract}
Traffic signal control plays a crucial role in urban mobility.
However, existing methods often struggle to generalize beyond their training environments to unseen scenarios with varying traffic dynamics.
We present TransferLight, a novel framework designed for robust generalization across road-networks, diverse traffic conditions and intersection geometries.
At its core, we propose a log-distance reward function, offering spatially-aware signal prioritization while remaining adaptable to varied lane configurations—overcoming the limitations of traditional pressure-based rewards.
Our hierarchical, heterogeneous, and directed graph neural network architecture effectively captures granular traffic dynamics, enabling transferability to arbitrary intersection layouts.
Using a decentralized multi-agent approach, global rewards, and novel state transition priors, we develop a single, weight-tied policy that scales zero-shot to any road network without re-training.
Through domain randomization during training, we additionally enhance generalization capabilities.
Experimental results validate TransferLight's superior performance in unseen scenarios, advancing practical, generalizable intelligent transportation systems to meet evolving urban traffic demands.
\end{abstract}

%

\section{Introduction}

Coordinating traffic at intersections is a major challenge for urban planning.
Due to the high and ever-increasing volume of traffic in city centres, intersections can
quickly become a bottleneck if traffic is not properly coordinated, which can lead to severe traffic congestion.
To avoid congested roads, signalized intersection are used to safely and efficiently coordinate traffic flows.
Traffic Signal Control (TSC) aims to optimise the traffic flow and related measures \cite{Wang2023}.

A common solution for TSC is to view it as an optimization problem by designing a mathematical model of the traffic environment using conventional traffic engineering theories and finding a closed-form solution based on that model. Provided that the assumptions inherent to the underlying traffic models are satisfied, such solutions produce good results in theory. However, assumptions such as uniform traffic \cite{Webster, Maxband, Roess2004} or unlimited vehicle storage capacity of lanes \cite{MaxPressure} are difficult or even impossible to observe in reality, which is why such solutions are not optimal in practice, especially when traffic demand is high and fluctuates significantly. Hence, the field pivoted towards adaptive signal control policies, which are learned from data through deep reinforcement learning (RL) \cite{Wei2021}. Yet, most existing works still struggle to effectively transfer their learned policies to changing traffic conditions.

\begin{figure}[t!]
    \centering
    \includegraphics[width=1.0\columnwidth]{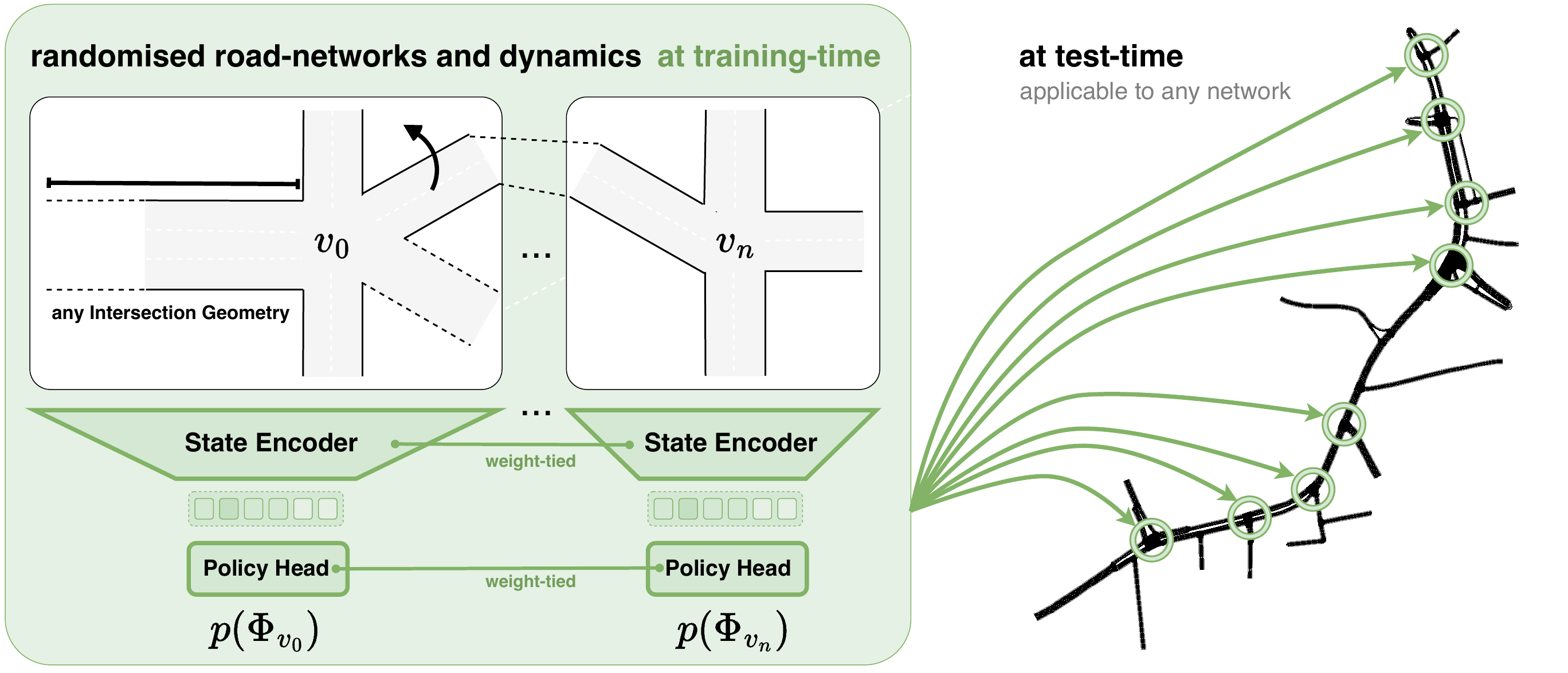}
    \caption{Our proposed traffic signal controller learns a general policy for flexible phase prediction during training. Due to the weight-tied models, we can apply the learned model to any road-network during inference.}
    \label{fig:pitch}
\end{figure}

\subsubsection{Rigid State and Action Spaces}
The majority of RL-based approaches employ overly rigid data structures to encode the mapping from states to actions.
Numerous studies simply encode states and actions as fixed-size vectors or spatial matrices \cite{Zheng2019, Wang2024}.
This approach inherently constrains the learned policy to a specific intersection geometry, which is defined by the structural arrangement of lanes, movements, and phases.
Consequently, the reusability of such models is limited to networks of homogeneous intersections with identical geometries \cite{IntelliLight, PressLight}.
In an attempt to increase flexibility, states and actions are (zero)-padded \cite{Zheng2019, MPLight}, introducing upper bounds to the system's diversity. However, due to the combinatorial explosion of possible intersection layouts, the required number of paddings grows exponentially \cite{Chu2019}, potentially compromising training efficiency and generalization ability of the model.

\subsubsection{Rigid Traffic Environments}
Another significant limitation in current RL-based approaches is the insufficient consideration of traffic dynamics' variability during training.
The majority of methods employ identical spatio-temporal traffic patterns across all training episodes \cite{IntelliLight, PressLight}.
While these models may exhibit impressive performance within this constrained setting, they typically suffer from substantial performance degradation when confronted with real-world variability \cite{Zheng2019, Yoon2021}.
This performance decline can be attributed to overfitting and the drastically constrained exploration space during training \cite{Jiang2024Explore}.
The limited exposure to diverse traffic scenarios during the learning phase results in models that lack robustness and adaptability to the complex and dynamic nature of real-world traffic conditions \cite{Korecki2023}.
Consequently, these models struggle to generalize effectively to the multifaceted and often unpredictable traffic patterns encountered in practical applications, highlighting a critical gap between laboratory performance and real-world efficacy.

\subsubsection{Degenerated Reward Functions}
Reinforcement Learning is driven by the choice of reward to be optimised.
As long-term objectives, like travel time, depend on a sequence of actions, credit assignment is difficult and might impact the training efficiency drastically.
Hence, short-term objectives are used instead, like waiting time or queue length \cite{Zheng2019, IG-RL, Devailly2024}, or weighted combinations of them \cite{IntelliLight, Yoon2021, Wu2023}.
Unfortunately, these rewards do not correlate, leading to different optima \cite{PressLight}.
As a solution, \citet{PressLight} showed that max-pressure control policy stabilise the traffic system over time, which lets queue length and travel time settle in a local optima.
Based on these guarantees, pressure-based rewards are frequently used in recent works \cite{AttendLight}.
However, as pressure is computed as a mean, it is invariant to various transformations of the input signal.
Different spatial locations of heavy traffic loads along the lane do not influence the indicator, leading to misjudgments of states.


\subsubsection{Towards General Control Policies}
The limitations of existing traffic signal control (TSC) approaches, particularly their inability to generalize across intersection and traffic conditions, necessitate a more robust and flexible solution.\footnote{The ideal solution would be a model capable of maintaining consistent, high-quality performance across a wide spectrum of road-networks and traffic dynamics. Once the general control policy is obtained, it can be applied to any (urban) environment.}
We present \emph{TransferLight}, a novel model that addresses these challenges by leveraging graph-structured representations and advanced training techniques.
Our contributions include 
\begin{itemize}
\item We introduce a novel log-distance reward function that provides a continuous, spatially-aware signal prioritizing near-intersection vehicles while remaining bounded and adaptable to diverse lane configurations, addressing key limitations of traditional pressure-based rewards \cite{PressLight}.
\item Building upon prior research \cite{Yoon2021, IG-RL, Devailly2024}, we propose a heterogeneous graph neural network \cite{Kipf2016} architecture for state encoding. This approach captures fine-grained traffic dynamics and enhances generalization, enabling universal applicability of the learned policy to varied intersection and road network geometries. 
\item We utilize domain randomization to vary both static and dynamic features of the traffic environment during the training process similar to \citet{IG-RL, Devailly2024}. This approach enhances the model's generalization capabilities to novel scenarios.
\item We use a decentralised multi-agent approach with a global reward and novel state transition priors to foster proactive decisions. This allows us to learn a single shared (weight-tied) general policy that can be zero-shot scaled to any road-network during test-time without re-training.
\end{itemize}
By combining these elements, \emph{TransferLight} overcomes the limitations of previous approaches, offering a unified framework for learning robust and adaptive traffic signal control policies.
Our experimental results demonstrate that \emph{TransferLight} achieves good performances on novel (unseen) scenarios, making a significant step towards practical, generalizable and intelligent transportation systems.

\section{Priliminaries}
\label{sec:prilimiaries}


\subsubsection{Traffic Signal Control}
We define a road network as a graph $G = (\mathcal{V}, \mathcal{I} \cup \mathcal{O})$, where
$\mathcal{V} = \{ v_k \mid k \in [1, 2 \ldots V]\}$ is the set of $V$ signalised junctions.
This geometric structure defines the environment for an agent to act on.
For notational convenience, we differentiate between incoming lanes $\mathcal{I}_v$ and outgoing lanes $\mathcal{O}_v$ for each intersection $v \in \mathcal{V}$.\footnote{Such that, $\mathcal{I} := \bigcup_{v \in \mathcal{V}} \mathcal{I}_v$ and $\mathcal{O} := \bigcup_{v \in \mathcal{V}} \mathcal{O}_v$.}
For situations, where we do not need to differentiate between incoming and outgoing lanes, we use $\ell \in \mathcal{I} \cup \mathcal{O}$ to denote an arbitrary lane.
Each lane $\ell$ defines a finite one-dimensional coordinate space $\ell \subset \mathbb{R}_+ \setminus \{0\}$ with its origin at the intersection's centre.

As in \cite{Wei2019Survey, NCHRP}, we define $m_v = (i, o)$ to be a movement from $i \in \mathcal{I}_v$ to $o \in \mathcal{O}_v$ with $m_v \in \mathcal{M}_v \subset \mathcal{I}_v \times \mathcal{O}_v$.
A movement can be either \emph{permitted}, \emph{prohibited}, or \emph{protected}. A movement is \emph{protected} if the associated road users have priority and do not have to give way to other movements. A movement is \emph{prohibited} if the signal is red, and it's \emph{permitted} if the associated road users must yield the right-of-way to the colliding traffic before they are allowed to cross the intersection.
A phase $\phi$ describes a timing procedure associated with the simultaneous operation
of one or more traffic movements \cite{NCHRP} with a green interval, a yellow change interval and an optional red clearance interval.
Let $\phi_v \in \Phi_v$ be a phase at intersection $v$, and $\mathcal{M}_{\phi_v} \subset \mathcal{M}$ be the associated right-of-way movements.
The phase set $\Phi_v$ defines the discrete action space for an agent acting on $v$.

This defines the static part of the environment.
The dynamics are given by a set of moving vehicles $\mathcal{C} = \{ c_k \mid k \in [1, 2 \ldots C]\}$.
These are modelled as points on the one-dimensional coordinate space $\ell$.
We define a state $\mathcal{C}_t$ by the vehicle positions at a time point $t$.
Hence, each vehicle's motion is captured by $c(t)$, which is evaluated at an a priori defined sampling frequency of the sensor (or the simulation).

\subsubsection{Cooperative Markov Games}
In a multi-intersection road network, agent coordination is crucial for efficient traffic flow.
This scenario extends the Markov Decision Process to a Markov Game \cite{Littmann1994}.
At each time step $t$, every agent $v \in \mathcal{V}$ observes the environment state $\mathcal{C}^t \in \mathcal{C}$ and selects an action $\phi_v^t \in \Phi_v$ using its policy $\pi_v(\phi_v^t \mid \mathcal{C}^t): \Phi_i \times \mathcal{C} \mapsto \mathbb{R}_+$.
The environment then transitions to $\mathcal{C}^{t+1}$ according to $T(\mathcal{C}^{t+1} \mid \phi^t, \mathcal{C}^t): \mathcal{C} \times \Phi \times \mathcal{C} \mapsto \mathbb{R}_+$, where $\Phi = \bigcup_{v \in \mathcal{V}} \Phi_v$ is the joint action space.
Each agent receives a reward $r_v^{t+1}$ based on $R_v(\mathcal{C}^t,\phi^t,\mathcal{C}^{t+1}) : \mathcal{C} \times \Phi \times \mathcal{C} \mapsto \mathbb{R}$, denoted as $R_v^t$ for brevity.

In fully cooperative Markov Games, the global reward is equivalent to individual rewards ($R^t = R_v^t, \forall v \in \mathcal{V}$) or a team average ($R^t = \frac{1}{|\mathcal{V}|} \sum_{v \in \mathcal{V}} R_v^t$).
While the former entails aligned goals for individual agents, the latter allows agents to pursue distinct objectives that contribute to the overall team benefit.
The objective is then to find a joint policy $\pi = \{ \pi_v \mid v \in \mathcal{V}\}$ that maximizes the expected discounted sum of global future rewards: 
\begin{equation} \label{eq:policy}
\pi^\ast = \argmax_\pi \mathbb{E}_{\mathcal{C}^t \sim \mu} \mathbb{E}\pi \left[ \sum_{k=0}^\infty \gamma^k R^{t+k} \mid \mathcal{C}^t \right],
\end{equation}
where $\mu(\mathcal{C}^t \mid \pi)$ is the stationary distribution of the Markov chain under joint policy $\pi$.\footnote{Note that \cref{eq:policy} is permutation invariant with respect to $\mathcal{V}$.
The geometric structure of the road network needs to be induced in the state representation.}

\section{Lifting Pressure-based Rewards}
\label{sec:reward}

Under mild assumptions\footnote{That is, no physical queue expansion for non-arterial environments and admissible average demand.}, a max-pressure control policy \emph{stabilises the traffic system} over time \cite{PressLight}. 
This means, that measures like queue length, throughput, and travel time settle in local optima.
We build upon these theoretical results by eliminating a remaining shortcoming of pressure-based systems.

\begin{figure}
    \centering
    \includegraphics[width=1.\columnwidth]{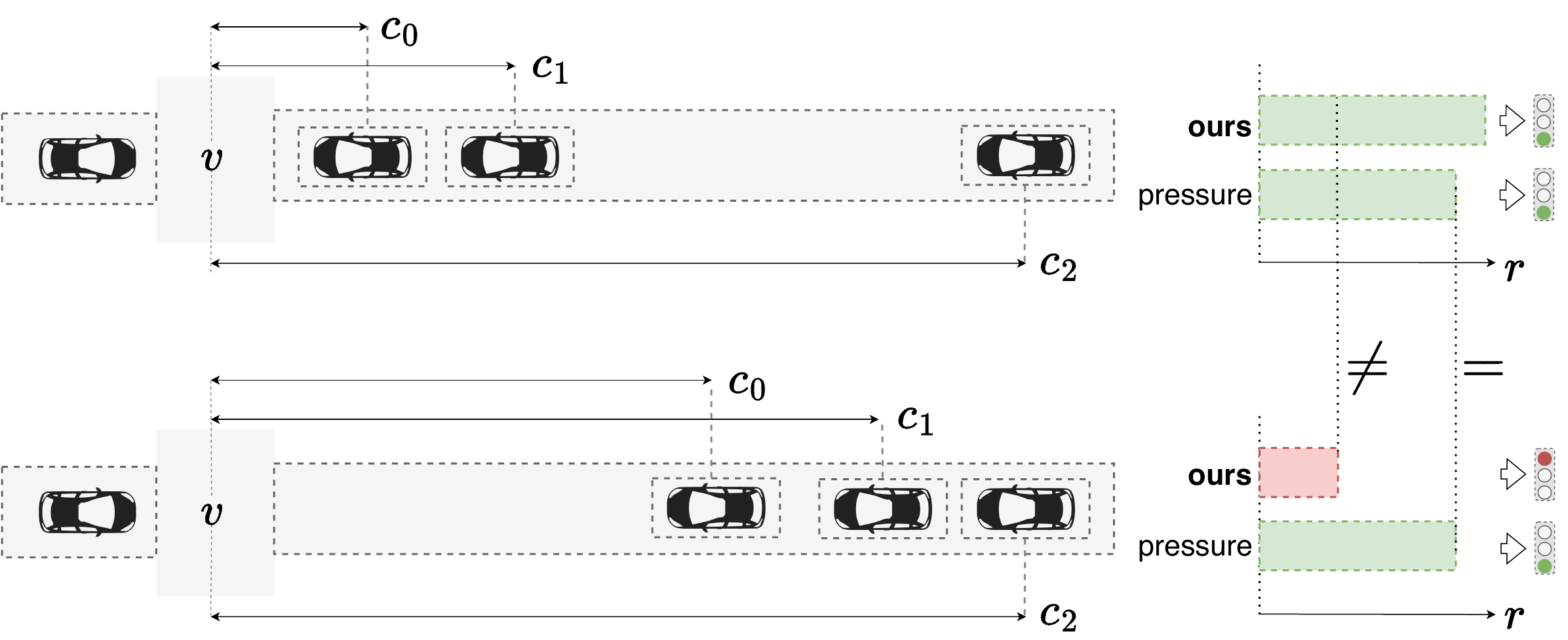}
    \caption{Pressure (see \cref{eq:pressure}) is symmetric to vehicle position translations within the lane's coordinate space.
    Our more expressive measure breaks this symmetry.}
    \label{fig:reward}
\end{figure}

\subsubsection{Degeneracies of Pressure}
\label{sec:degeneracies}

We prove that the pressure of a movement suffers from several degeneracies introducing plateaus to the reward surface, which prohibit convergence to superior extrema. 
The pressure $\rho(m)$ of a movement $m \in \mathcal{M}$ \cite{PressLight} is defined by the difference between the incoming and outgoing vehicle densities, such that
\begin{equation} \label{eq:pressure}
    \frac{C_i}{|i|} - \frac{C_o}{|o|} \quad \text{with} \quad m = (i,o),
\end{equation}
where $|i|, |o| \in \mathbb{R}_+ \setminus \{0\}$ are the length of the lanes.
Densities are computed by the arithmetic mean over vehicles \footnote{We can interpret the traffic density as $\frac{1}{|\ell|} \sum_{p \in \ell} \mathds{1}_p$, where $\mathds{1}_p \in \{0,1\}$ is an indicator returning $1$ if there is a vehicle at the spatial $p$ on $\ell$.}, which comes with the following fundamental properties arsing from the linearity of the operation:
\begin{itemize}
    \item \emph{permutation invariance}, $\operatorname{mean}(\mathbf{x}) = \operatorname{mean}(\pi\mathbf{x})$,
    \item \emph{translation equivariance}, $\operatorname{mean}(\mathbf{x} + b) = \operatorname{mean}(\mathbf{x}) + b$,
    \item \emph{scale equivariance}, $\operatorname{mean}(b\mathbf{x}) = b\operatorname{mean}(\mathbf{x})$,
\end{itemize}
for any sequence $\mathbf{x} \in \mathbb{R}^n$, permutation matrix $\pi \in \{0,1\}^{n \times n}$, and $b \in \mathbb{R}$.
These symmetries also apply locally, such that $\operatorname{mean}(\{ x_1 + b, x_2 - b, \ldots, x_n\}) = \operatorname{mean}(\{ x_1, x_2, \ldots, x_n\})$ for instance.
By these relations, \emph{equivalence classes} are formed, i.e., subsets with constant outputs under these transformed inputs.
Hence, the pressure stays constant, when permuting the positions of vehicles, shifting vehicles along the lane or scaling the distribution of vehicles.
The latter two are of specific interest, as the first one would not change the state $\mathcal{C}_t$.

Modelling the reward function by (pure) pressure maps these equivalence classes on the reward surface and with that on the loss surface.
As gradients on these plateaus are exactly zero, gradient-based optimisation will fail leaving these regions.
This might be mitigated by a drastically increased momentum term \cite{Adam}, allowing the model to jump over these regions.
However, the model can extract valuable information from these regions \emph{iff} these degeneracies are lifted.

\subsubsection{Lifting the Degeneracies}
\label{sec:lifting}

We argue, that the degeneracies of \cref{eq:pressure} can get lifted by inducing spatial information.
As stated in \cref{sec:prilimiaries}, every vehicle $c$ can be interpreted as a point on the lane's one-dimensional coordinate space $\ell$.
Using the Euclidean distances does not lift the degeneracies\footnote{For example, a configuration with a single vehicle at a large distance from the intersection's centre would yield the same metric value as a configuration with multiple vehicles positioned closer to the centre, provided the sum of their distances is equal to that of the single distant vehicle.}
Instead, we use $\log$-distance, defined by $\log \left( c + \epsilon \right) \in [\log \epsilon \approx -\infty, \log (1+\epsilon) \approx 0]$, where $\epsilon \approx 0$. 
The farther away a vehicle $c$ from $v$, the larger the $\log$-distance (closing in on $0$).
This can be computed for an entire lane $\ell \in \mathcal{I} \cup \mathcal{O}$ by $\left\{ \log c + \epsilon \mid c \in \mathcal{C}_\ell \right\} = \log \left( \mathcal{C}_\ell + \epsilon \right)$.

We interpret the cumulated $\log$-distances as the negative energy of the system.
Analogously to a simplified potential energy of a system of particles, where the energy increases with distance between particles, as leveraged in \cite{Schmidt2024}.
\emph{The goal is to minimise this energy, i.e., push the densities away from the intersection's centre.}
We interpret the total $\log$-distance as the energy $E_\ell \in \mathbb{R}_+$ of the lane,
\begin{equation}
    \hat{E}_\ell = \sum_{c \in \mathcal{C}_\ell} \log \left( c + \epsilon \right).
\end{equation}
This breaks both the translation and scale equivariance.\footnote{This follows from $\log \left( c + \epsilon + b \right) \neq b + \log \left( c + \epsilon \right)$ for $b\neq0$ and $\log \left( bc + \epsilon \right) \neq b \log \left( c + \epsilon \right)$.}
Therefore, $\hat{E}$ lifts the degeneracies of $E$ using the symmetry-breaking $\log$-distance formulation.
We define the cumulated vehicle positions on a lane to be its energy, $E_\ell := C_\ell \in \mathbb{R}_+$.
With this, we can formulate the average $\log$-pressure by the cumulated and normalised $\log$-distances,
\begin{equation}
    \sum_{(i, o) \in \mathcal{M}_v} \frac{1}{|i|}\hat{E}_i - \frac{1}{|o|}\hat{E}_o.
\end{equation}
With this we define the reward
\begin{equation} \label{eq:log_distance_reward}
    r_i = - \left| \sum_{(i, o) \in \mathcal{M}_v} \frac{1}{|i|}\hat{E}_i - \frac{1}{|o|}\hat{E}_o \right|.
\end{equation}
We focus on cooperative Markov Games \cite{Littmann1994}, where agents have an incentive to work together to achieve a team goal, which can be expressed by a global reward function $R(t)$.
In such Multi-Agent settings, sharing information among agents is key, as the other agents induce otherwise unpredictable dynamics (non-stationary environments), which limits cooperation \cite{Zhang2020}.
This can be done by joint state and action spaces, which, however, require supportive mechanisms to cope with the exponentially growing joint spaces \cite{Choudhury2021}.
Hence, action and state spaces are often disjoint and agents are trained by a global reward function to encourage cooperation \cite{PressLight, MPLight, Pol2022}.
In the following, we propose our state encoding to cope with these challenges.

\section{Graph-Structured State Encoding}
\label{sec:state_space}

\begin{figure}
    \centering
    \includegraphics[width=1.\columnwidth]{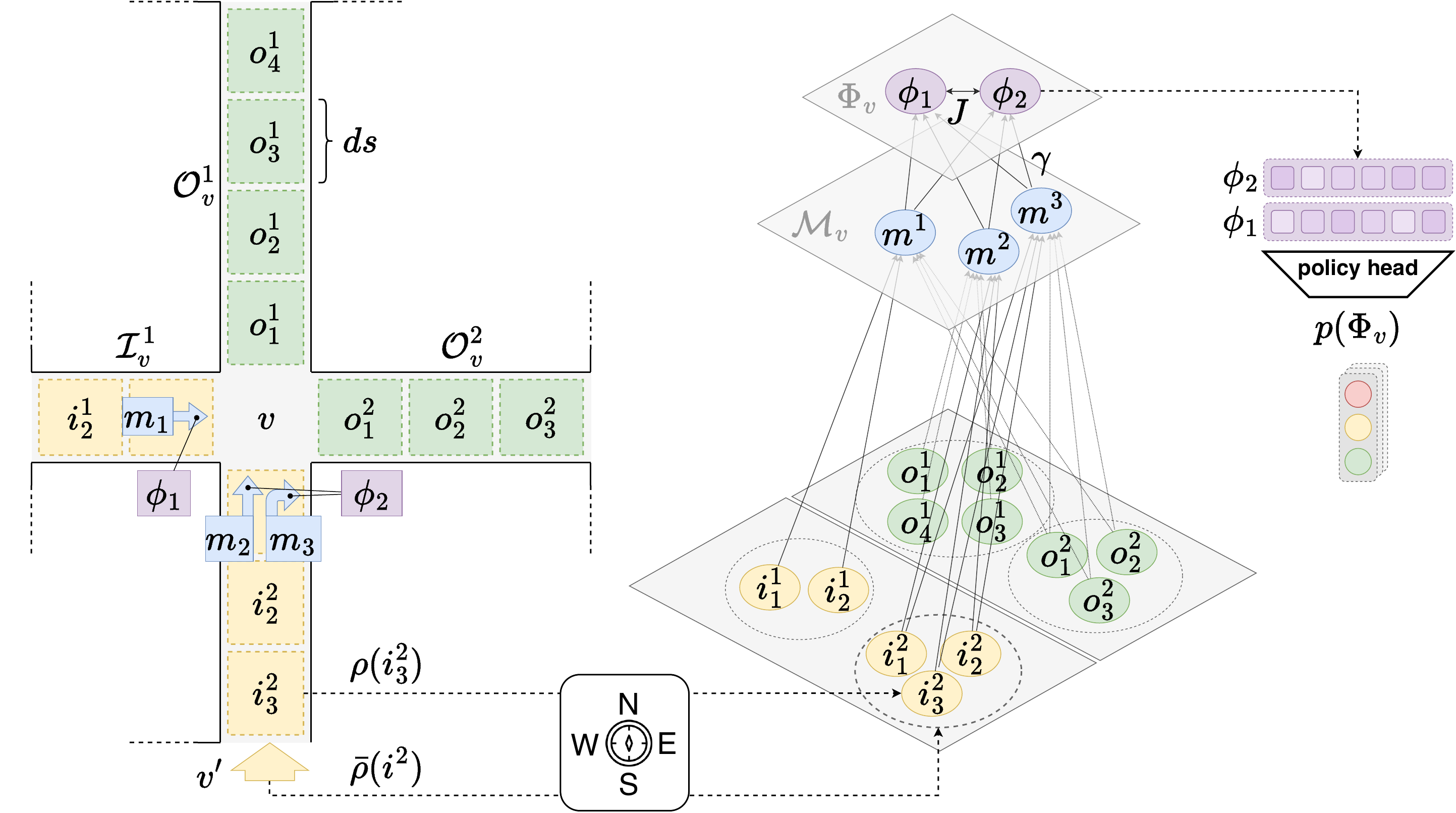}
    \caption{Our hierarchical state space encoding uses a position-encoded segment-density set on the lowest level.
    This information is embedded and aggregated to form movement representations, which then undergo another pass to the phase level.
    On the phase level, we have intra-level updates, otherwise information are passed down-to-top along the directed heterogenous graph structure.}
    \label{fig:state}
\end{figure}

Following \citet{IG-RL, Devailly2024}, we utilize a graph neural network on a heterogeneous graph to encode both static and dynamic state characteristics of individual intersections.
\emph{This allows us to encode any intersection geometry} regardless of the length of lanes and the number of lanes, approaches, movements and phases.
By sharing the parameters across all intersections in the network, \emph{the model is encouraged to converge to a policy that generalises various intersection configurations and traffic conditions}.
We contextualise encodings by state transition priors to \emph{allow for proactive decisions} (which enable green waves).
We provide an illustration of our state encoding in \cref{fig:state}.
We will discuss its core elements in the following.

\subsubsection{Lane Partitioning}
\label{sec:segmentation}

As stated in \cref{sec:lifting}, the density estimate over $\ell$ suffers from degeneracies.
A state encoder using these estimates as inputs would inherit the degeneracies, which would smooth out update nuances.
Instead, we bound the degeneracies to only act in limited sub-spaces.
We define a metric $ds$ to partition $\ell$ into $\frac{\ell}{ds}$ equally-sized segments, which defines a hyperparameter to control the resolution of the measure applied on top of lanes.
In each segment, we estimate the density by $\rho(s) = \frac{1}{ds} C_s$.
This factors out the number of vehicles for the input to the encoder (similar to the density estimate over $\ell$).
Such a representation ensures that the dynamical traffic system is fully described (for a proof refer to \citet{PressLight}).

While often overlooked in previous studies \cite{Zheng2019, AttendLight, Metalight, Yoon2021}, the length of lanes or segments plays a crucial role in traffic dynamics.
Our approach employs a uniform and constant segment length $ds$, thereby streamlining the input feature set compared to related works \cite{IG-RL, Devailly2024}.
This design choice allows the policy to implicitly learn length-related characteristics, including segment capacity, enhancing the model's adaptability to diverse road networks compared to prior work  \cite{PressLight}.
However, $ds$ has to be small enough to minimise the impact of local degeneracies, as discussed in \cref{sec:degeneracies}.


\subsubsection{Transition Prior}
\label{sec:prior}

Modelling only the dynamics within the boundaries of the intersection, would result in reactivity rather than proactivity, especially when $\ell$ is small.
To fix this, we interpret the road-network as a coordination graph, which allows us to induce additional context to each agent $v \in \mathcal{V}$.
We define the connectivity of the coordination graph by movements,
\begin{align} \label{eq:prior_movements}
    \mathcal{M}_{\leftarrow \ell} &:= \{ (\ell, o^\prime) \mid i^\prime = \ell; (i^\prime, o^\prime) \in \mathcal{M}\}, \nonumber \\
    \mathcal{M}_{\rightarrow \ell} &:= \{ (i^\prime, \ell) \mid \ell = o^\prime; (i^\prime, o^\prime) \in \mathcal{M}\}.
\end{align}
This gives a single-hop receptive field for every $v \in \mathcal{V}$.
This locally interdependent structure \cite{Yi2024} can be interpreted as modelling communication channels between $v$ and its adjacent neighbour intersections.\footnote{Agents are incentivized to cooperate rather than act solely in their self-interest. This can lead to more stable equilibria where multiple agents coordinate their strategies effectively.}
We use this to define a state transition prior
\begin{equation} \label{eq:prior}
    \bar{\rho}_\ell = \sum_{(i,o) \in \mathcal{M}_{\rightarrow \ell}} \rho(i_0) - \sum_{(i,o) \in \mathcal{M}_{\leftarrow \ell}} \rho(o_0),
\end{equation}
where $i_0$ and $o_0$ are the closest segments to the intersection's centre.
If $\bar{\rho}_\ell < 0$, more vehicles are going to leave $\ell$.

\subsubsection{Lane Coordinate Frames}
\label{sec:lane_coord_frame}

To break the permutation invariance of the segment set, we define the centre of the intersection as a reference point and induce a positional encoding on the segments relative to that point.
We use sinusoidal positional encoding \cite{Vaswani2017} along segments on each lane and over lanes.
Instead of additive fusion \cite{Vaswani2017}, we concatenate the positional information with the density of the segment.
This preserves both identities, which improves expressiveness without the need of separate processing \cite{Yu2023}.
Thus, we can define the segment feature vector by $\mathbf{h}_s = [\rho(s) \| \operatorname{pe}(s) \| \bar{\rho}_\ell]^\top \in \mathbb{R}^d$ be the feature vector of a segment $s \in \ell$. 

\subsubsection{Segment-to-Movement Encoding}
\label{sec:seg2mov}

We apply a graph attention network \cite{Velickovic2017} to learn the mapping $\mathbb{R}^{\frac{\ell}{ds} \times d} \mapsto \mathbb{R}^{d^\prime}$.
To improve expressiveness, we use dynamic scoring \cite{Brody2022} to compute attention weights
\begin{equation} \label{eq:attention}
    \alpha_{s} = \frac{\exp u(\mathbf{h}_s)}{\sum_{s^\prime \in \mathcal{N}_\ell} \exp u(\mathbf{h}_{s^\prime})}
    \quad \text{with} \quad
    u(\mathbf{h}_s) = \mathbf{a}_s^\top \sigma \left( \mathbf{W}_s \mathbf{h}_s \right),
\end{equation}
where $\mathbf{a}_s \in \mathbb{R}^{d^\prime}$ and $\mathbf{W}_s \in \mathbb{R}^{d \times d^\prime}$ are learnable weights.
$\mathcal{N}_\ell$ defines the segment set for lane $\ell$ and $\sigma$ is a monotonic non-linearity, like Leaky-ReLU. 
We then compute a representation for each movement $\mathbf{h}_m \in \mathbb{R}^{d^\prime}$ by a weighted average of its segments, such that
\begin{equation} \label{eq:aggregation}
    \mathbf{h}_m = \sigma \left( \mathbf{b}_s + \underbrace{\hat{\mathbf{W}}_s \mathbf{h}_s}_{\text{residual}} + \sum_{s \in \mathcal{N}_i} \alpha_{s} \mathbf{W}_s \mathbf{h}_s + \sum_{s \in \mathcal{N}_o} \alpha_{s} \mathbf{W}_s \mathbf{h}_s \right),
\end{equation}
where $\hat{\mathbf{W}}_s \in \mathbb{R}^{d \times d^\prime}$ enables learnable residual connections and $\mathbf{b}_s \in \mathbb{R}^{d^\prime}$ being the bias term.
Movement nodes do not hold information initially, hence the update is independent of the original target node features $\mathbf{h}_m$.\footnote{$\mathbf{h}_m$ is initialised with zeros, neutralising its impact in \cref{eq:aggregation}.}
To ensure that the neighbourhood aggregation runs in a numerically stable manner while allowing for a high degree of representational strength, the individual aggregation functions are implemented as weighted sums with multi-head attention.
We compute attention over incoming and outgoing segments separately, but aggregate and update the movement node features in parallel.

\begin{figure*}
    \centering
    \includegraphics[width=1.\textwidth]{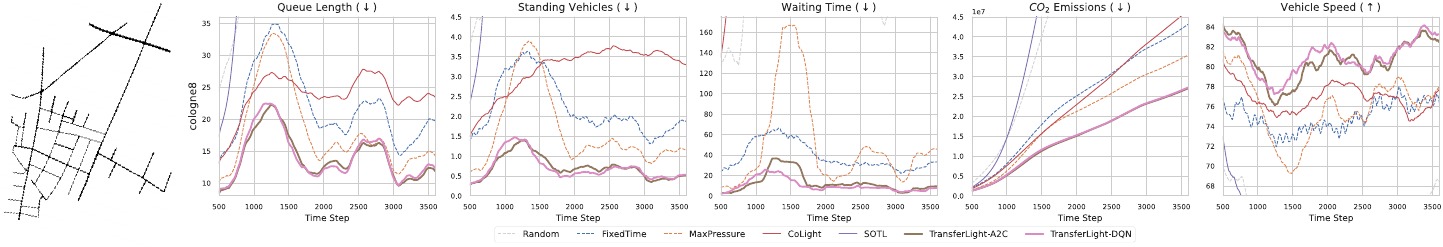}
    \caption{Test performances (moving averages) on \emph{Cologne8} over 3600 simulated time steps.}
    \label{fig:genalised}
\end{figure*}

\subsubsection{Movement-to-Phase Encoding}
\label{sec:mov2pha}

The obtained movement node features $\{ \mathbf{h}_m \mid m \in \mathcal{M}_v\}$ form another heterogeneous directed acyclic sub-graph with the phase nodes.
Instead of a sparsified graph, we use a fully-connected bipartite structure with additional edge features.
Each connection between a movement $m \in \mathcal{M}_v$ and a phase $\phi \in \Phi_v$ holds a scalar $\gamma_{m\phi} \in \{ -1,0,1 \}$ as an edge feature indicating whether a movement is prohibited, protected or permitted during a phase.
In literature, often only permitted, or protected movements are considered \cite{Zheng2019, Metalight}.
We argue, that also the information about prohibited movements are essential to determine the energy of a phase.
Furthermore, phase nodes are initialised by a binary flag $\mathbf{h}_\phi \in \{ 0,1 \}$ indicating whether the phase is currently active or not.
This changes \cref{eq:attention} and \cref{eq:aggregation} to
\begin{align}
    u(\mathbf{h}_{m\phi}) &= \mathbf{a}_m^\top \sigma \left( \mathbf{W}_m \mathbf{h}_m + \mathbf{W}_\phi \mathbf{h}_\phi + \mathbf{W}_{\gamma} \gamma_{m\phi} \right) \\
    \text{and} \quad \mathbf{h}_\phi &= \sigma \left( \mathbf{b}_m + \tilde{\mathbf{W}}_m \mathbf{h}_m + \sum_{m \in \mathcal{M}_v} \alpha_{m\phi} \mathbf{W}_m \mathbf{h}_m \right), \nonumber
\end{align}
where $\mathbf{a}_{m\phi}, \mathbf{b}_m \in \mathbb{R}^{d^\prime}$ and $\mathbf{W}_m, \tilde{\mathbf{W}}_m \in \mathbb{R}^{d^\prime \times d^\prime}$ are learnable weights.
Attention weights are computed as in \cref{eq:attention} but normalised over $\mathcal{M}_v$ instead.
Contrary to \citet{Velickovic2017}, we embed node and edge features separately, which reduces the model complexity while still preserving expressiveness.
Furthermore, we use the edge features and the initial phase flag only to compute the attention scores.
Hence, the model can use $\gamma$ to weight movement features during aggregation, but it does not make any further inferences from the movement information.
As we use a directed acyclic graph, we do not face the identity issue discussed in general edge-based graph attention \cite{Wang2021}.
This form of aggregation also preserves permutation invariance.
In contrast to the level before, this is an important property for the encoding of phases, as they should be orientation independent \cite{Zheng2019}.
The obtained phase node representations are further leveraged in an intra-level propagation phase, as discussed next.

\subsubsection{Intra-Level Phase Propagation}
\label{sec:pha2pha}

We model the connection between phases as a fully-connected homogeneous graph  with Jaccard coefficients $J_{\phi \phi^\prime} \in \mathbb{R}_+$ between each phase pair $\phi, \phi^\prime \in \Phi_v$.
The Jaccard coefficient encodes the intersection over the union of the green signals between the two phases.
This structures the phase space by quantifying the relative differences between phases w.r.t. to their “green” portions. 
This results in the following intra-level update formulation
\begin{align}
    u(\mathbf{h}_{\phi \phi^\prime}) &= \mathbf{a}_\phi^\top \sigma \left( \mathbf{W}_\phi \mathbf{h}_\phi + \mathbf{W}_\phi \mathbf{h}_{\phi^\prime} + \mathbf{W}_J J_{\phi \phi^\prime} \right) \\
    \text{and} \quad \mathbf{h}_\phi &\leftarrow \sigma \left( \mathbf{b}_\phi + \tilde{\mathbf{W}}_\phi \mathbf{h}_\phi + \sum_{\phi \in \Phi_v} \alpha_{\phi \phi^\prime} \mathbf{W}_\phi \mathbf{h}_\phi \right), \nonumber
\end{align}
where $\mathbf{a}_\phi, \mathbf{b}_\phi \in \mathbb{R}^{d^\prime}$ and $\mathbf{W}_\phi, \tilde{\mathbf{W}}_\phi \in \mathbb{R}^{d^\prime \times d^\prime}$ are learnable weights.
Again, attention weights are computed as in \cref{eq:attention} but normalised over $\Phi_v$ instead.
After propagation, each node holds weighted information about all other phases, which renders a single layer sufficient. 

\subsubsection{Weight-Sharing}
\label{sec:weight_sharing}

Our universal state encoding function allows using the model for each intersection.
In this way, \emph{our model can be applied to any road-network size.}
Moreover, by sharing parameters among agents, the algorithms are essentially encouraged to converge to a region in parameter space that works well for arbitrary intersections and traffic conditions, thereby promoting generalization.



\section{Domain-Randomised Training}
\label{sec:training_generator}

Domain Randomization (DR) is a powerful technique for bridging the sim-to-real gap \cite{Tobin2017}. By introducing sufficient variability in the simulated source domain during training, DR enables the agent to generalize its policy to the target domain, treating it as another variant within its learned distribution.
The core principle of DR involves configuring the environment based on a randomly sampled configuration $\xi \sim \Xi$, where $\Xi$ represents the space of possible domain parameters.
$\Xi$ contains all traffic-networks under some degree of freedom, as well as different forms of traffic dynamics.
The agent's objective is to find an optimal policy $\pi^\ast$ that maximizes the expected return across all possible environmental configurations, i.e., extending \cref{eq:policy}
\begin{equation}
    \pi^\ast = \argmax_\pi \mathbb{E}_{\xi} \mathbb{E}_{\mathcal{C}^t} \mathbb{E}_\pi \left[ \sum_{k=0}^\infty \gamma^k R(t+k) \mid \mathcal{C}^t, \xi \right],
\end{equation}
where $\mathcal{C}^t \sim \mu(\mathcal{C}^t \mid \pi; \xi)$ denotes the stationary distribution of the Markov chain under configuration $\xi$ and policy $\pi$.
We sample the static environmental characteristics (like the number of intersections and lane lengths) from a uniform distribution a priori.
For the dynamics, we use traffic flow modelling to define each flow $f \in \mathcal{F}$ by its route, vehicle count, and departure times.
To enhance realism and variability, we model departure times using a beta distribution with flow-specific parameters:
\begin{equation}
    \mathcal{T}^f = \{ t_\text{max} b_k \mid b_k \sim \operatorname{Beta}(\alpha^f, \beta^f), 1 \leq k \leq C^f\},
\end{equation}
where $\alpha^f, \beta^f$ are sampled from a uniform distribution.\footnote{We sample a destination and target line segment and use the Dijkstra algorithm \cite{Dijkstra} to estimate the shortest path. We use $\alpha^f, \beta^f \sim \operatorname{Unif}(1, 10)$ in our experiments. The number of vehicles following the flow is sampled from a pool of $C$ available vehicles in the simulation.}
In literature, a Poisson process with a constant rate of $\frac{C^f}{t_\text{max}}$ vehicles per second with $t \in [0, t_\text{max}]$ is often used instead.
However, the constant departure rates are often not realistic in practice (e.g. during rush hours).
This approach allows for diverse departure patterns, including peaks and fluctuations, while still encompassing the possibility of constant departure rates.


\section{Experiments} \label{sec:experiments}

The primary objective of our experiments is to show the ability of \emph{TransferLight} to transfer its control policy to novel scenarios without requiring any kind of re-training or fine-tuning.
In all experiments, \emph{TransferLight} is trained on randomly generated road-networks with random traffic dynamics and tested on a yet unseen benchmark.
This allows us to quantify the generalisability of our method explicitly.
In \cref{sec:generalisation}, we analysed various performance measures on multiple benchmarks (test scenarios) with several well-known baselines.
As arterial scenarios are of specific interest for the community \cite{PressLight}, we conduced a detailed investigation of our model's generalisability on such scenario types (see \cref{sec:signal_progression}).
The software specifications of our implementations can be found in our open-sourced code\footnote{\texttt{https://github.com/johSchm/TransferLight}}.

\paragraph{Exchangeable Policy Heads}
The learnable hierarchical state encoding \cref{sec:state_space} maps states to action (phase) energies.
The policy control function maps from this action energy space to action probabilities.
This results in maximum flexibility when it comes to the policy function.
In this work, we chose a Double DQN \cite{DoubleDQN} and a A2C \cite{A2C} as policy heads, but any other can be used instead.




\subsection{Generalising different Scales}
\label{sec:generalisation}

\begin{table}[htbp]
\centering
\small
\caption{Average number of standing vehicles ($\downarrow$) over 3600 simulated time steps (TL = TransferLight).}
\label{tab:single_intersection}
\begin{tabular}{lcc}
& \emph{Cologne1} & \emph{Ingolstadt1} \\
& \includegraphics[angle=90, width=0.15\textwidth]{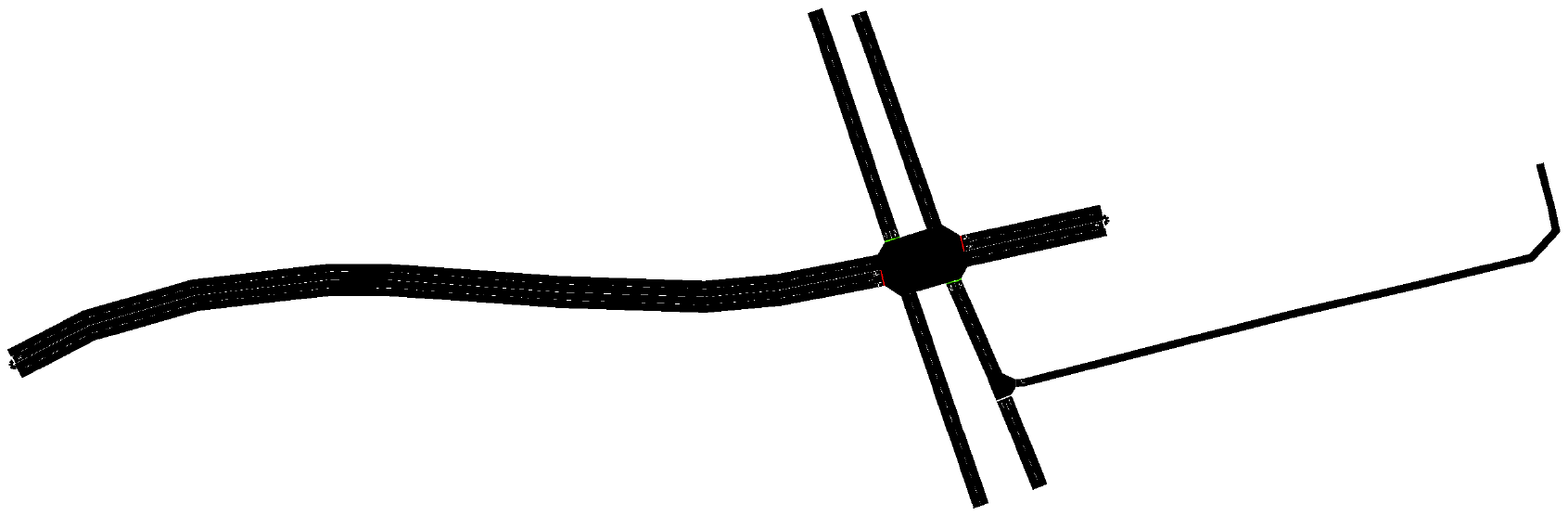} & \includegraphics[angle=90, width=0.15\textwidth]{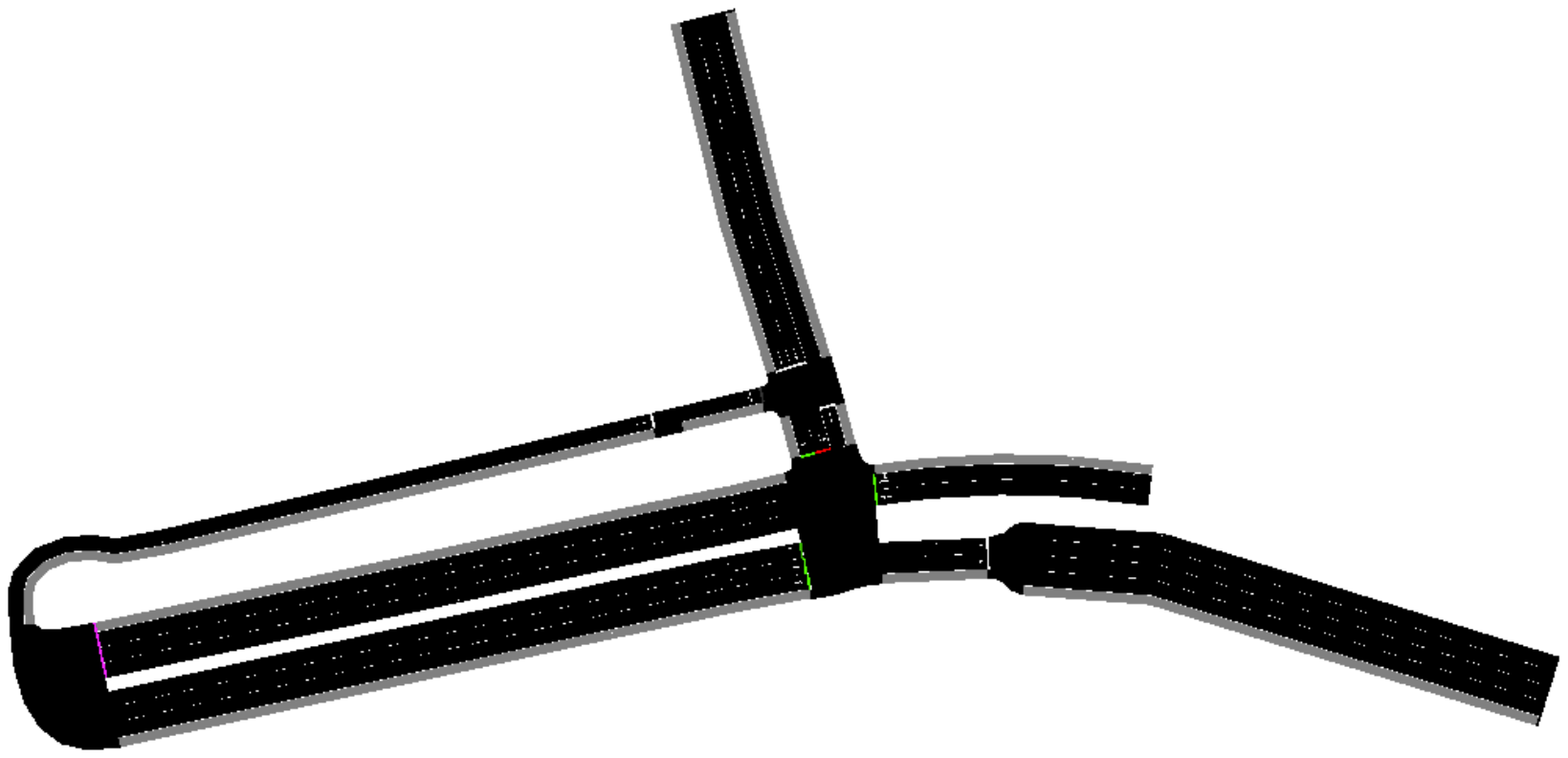} \\
Random & 40.90 \small{$\pm 21.77$} & 8.41 \small{$\pm 6.34$} \\
FixedTime & 14.58 \small{$\pm 8.37$} & 7.04 \small{$\pm 6.95$} \\
MaxPressure & 8.00 \small{$\pm 5.22$} & 1.88 \small{$\pm 1.38$} \\
\emph{TL-DQN} & 6.70 \small{$\pm 5.72$} & 1.93 \small{$\pm 2.08$} \\
\emph{TL-A2C} & 7.21 \small{$\pm 7.04$} & 2.30 \small{$\pm 3.30$} \\
\end{tabular}
\end{table}

A general traffic signal control policy should be able to generalise from single intersections to more complex road networks.
We demonstrate this ability by conducting experiments on either end.
\Cref{tab:single_intersection} compares our models to different baselines on two single-intersection benchmarks.
We analysed the number of vehicles, as for a single intersection this measure seems the most reasonable.
We found that both \emph{TransferLight} variants outperform all baselines on \emph{Cologne1} and perform quasi on par with MaxPressure, causing the least congestion.

To analyse how are policy scales to more complex road networks, we conducted an experiment on \emph{Cologne8} comprising 8 signalised intersections.
We measured multiple popular traffic performance indicators during testing.
We found that both \emph{TransferLight} variants outperformed all heuristic and trained baselines. 
Note that, CoLight \cite{CoLight} and SOTL \cite{SOTL} are explicitly trained on \emph{Cologne8}, whereas \emph{TransferLight} generalises from random road-networks.
Both trained baselines failed to control a subset of intersections, leading to early congestions and hence the worse performance.
The results in \cref{fig:genalised} undermine the ability of \emph{TransferLight} to generalise also to more complex scenarios.
In the appendix, we rise the problem complexity even more to identify \emph{TransferLight's} generalisation limits.

\subsection{Arterial Signal Progression}
\label{sec:signal_progression}

\begin{figure}
    \centering
    \includegraphics[width=1.\columnwidth]{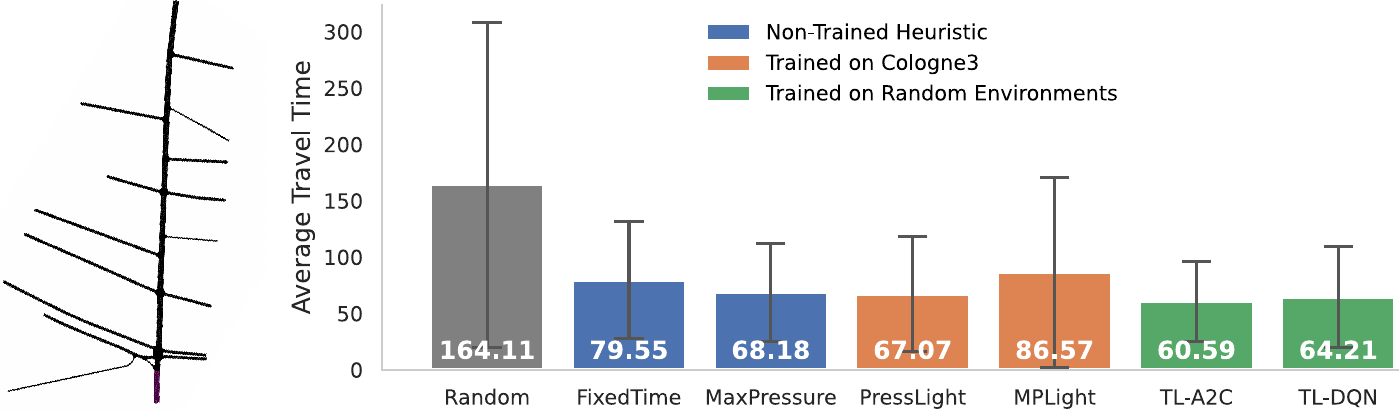}
    \caption{Average Travel Time on \emph{Cologne3} over 3600 simulated time steps.}
    \label{fig:cologne3_experiment}
\end{figure}

\begin{figure}
    \centering
    \includegraphics[width=1.\columnwidth]{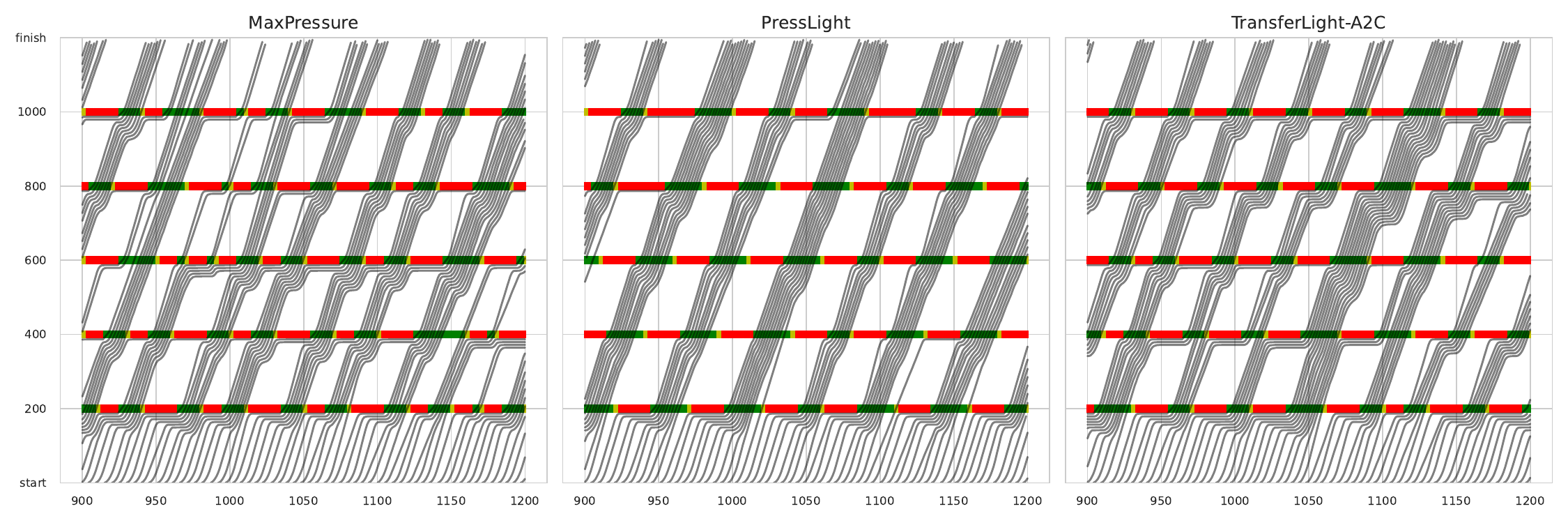}
    \caption{Signal progression comparison on a synthetic 5-intersection arterial scenario. PressLight \cite{PressLight} was explicitly trained and designed to fit this specific scenario, whereas \emph{TransferLight} generalises from random non-arterial road-networks.}
    \label{fig:progression}
\end{figure}

A special type of coordination is signal progression, which attempts to coordinate the onset of green times of
successive intersections along an arterial street in order to move road users through
the major roadway as efficiently as possible \cite{Wang2023}.
Intuitively, the hope here is to create a green wave in which green times are cascaded so that a large group of vehicles (also
called a platoon) can pass through the arterial street without stopping.

PressLight \cite{PressLight} and MaxPressure were shown to maximise throughput and minimise travel time in arterial environments.
We compare \emph{TransferLight} to these baselines while \emph{not} being trained on arterial scenarios (other than PressLight).
Here, the state transition prior is essential to provide geometric information to perform proactive decisions.
\Cref{fig:progression} shows the spatio-temporal signal progression plots, where each gray line represents the trajectory of a single vehicle.
In the optimal case, vehicle trajectories form straight lines (i.e., they keep a constant velocity).
We found that the zero-shot performance of our model can keep up with the performances of the baselines.
In \cref{fig:cologne3_experiment}, we extended the experiment to a real-world scenario.
We found that \emph{TransferLight} was able to achieve the minimal travel time among the contesters, including MPLight \cite{MPLight} and PressLight.
Our model learns a more robust and general policy from the DR-based training, enhancing its effectiveness in real-world environments characterized by greater variability.

\section{Conclusion}

We presented a novel framework designed for robust generalization across road-networks, diverse traffic conditions and intersection geometries.
Our method can scale to any road-network through a decentralized multi-agent approach with global rewards and state transition priors to ensure proactive decisions.
We used a heterogeneous and directed graph neural network to encode any intersection geometry, which we train using a novel log-distance reward function.
Generalization is further fostered by domain randomization during training.
This is particularly valuable for real-world applications where traffic conditions can vary significantly due to events, road closures, or long-term changes in urban mobility patterns.


\subsubsection{Limitations and Future Work}
Our method shows already striking generalisation capabilities, which, however, need further improvement to cope with even larger road networks. 
In future work, we aim to extend the concept of symmetry breaking to the intersection's geometries.
Mapping intersections to canonical forms, as in \citet{Jiang2024}, collapses the state space to an exponentially smaller subspace.
These canonical forms can be obtained from equivariant encodings \cite{Pol2022} using canonicalisation priors \cite{Kaba2023, Mondal2023} or by search \cite{Schmidt2024ITS}. 
This will drastically improve the sample efficiency of our model and render domain randomisation useless.

\section{Acknowledgments}
We would like to thank the Thorsis Innovation GmbH and Galileo Test-Track team for valuable support throughout this work.
Furthermore, the authors acknowledge the financial support by the Federal Ministry of Education and Research of Germany (BMBF) within the framework for the funding for the project PASCAL. 

\bibliography{library}

\newpage
\appendix
\section{Supplementary Material}

\begin{figure}
    \centering
    \includegraphics[width=1.\columnwidth]{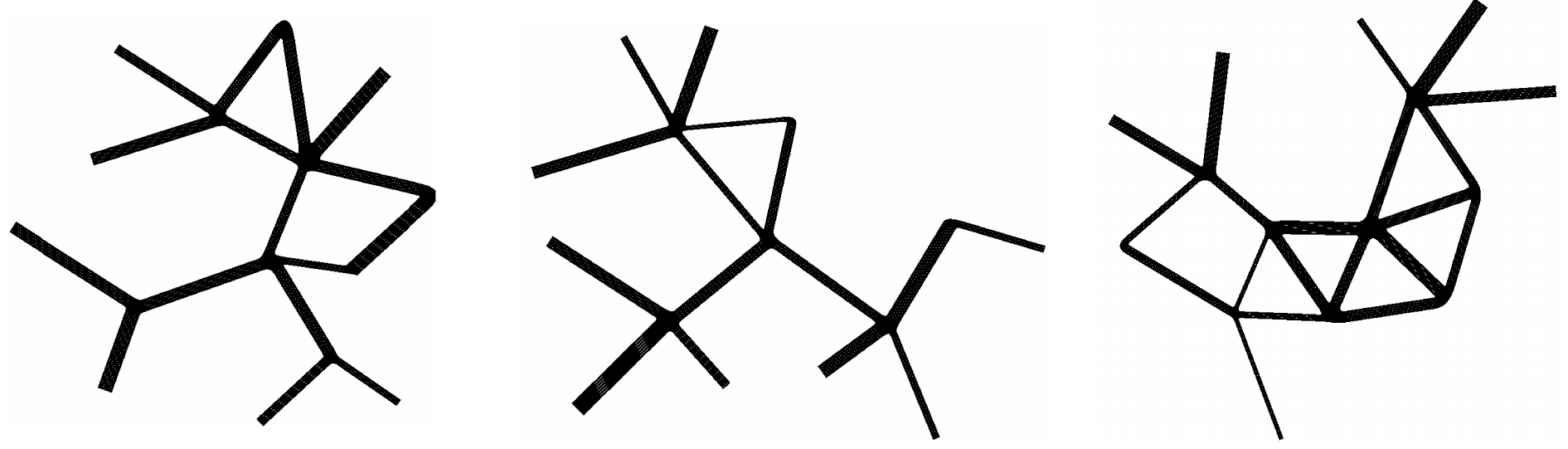}
    \caption{Three random road-network samples (static environments) used during training.}
    \label{fig:random_road_nets}
\end{figure}

\subsection{Implementation Details}
A replay buffer is introduced to decorrelate the experience tuples used for updating the parameters of the online DQN.
For the A2C tuples are promptly utilized to perform immediate updates.
This immediacy is crucial, as estimating the policy gradient necessitates the use of experience tuples generated from the current policy.
All learnable functions are MLPs incorporating additional intermediate layers for layer normalization and dropout.
This design aims to enhance training stability and convergence.
We used a $64$-dimensional latent space, which is significantly smaller than all our baselines, saving computational resources, allowing for better scalability and faster inference.
We used $8$ attention heads for all attention-based graph layer (see \cref{sec:state_space}).
All experiments are performed on an Nvidia A40 GPU (48GB) node with 1 TB RAM, 2x 24core AMD EPYC 74F3 CPU @ 3.20GHz, and a local SSD (NVMe).
As the inference costs are generally extremely cheap, the available resources are only required to amplify training.
More details can be found in our open-sourced code base.

\subsubsection{Simulation Details}
We used the SUMO (Simulation of Urban MObility) \cite{SUMO} during all our experiments.
As in \cite{PressLight}, each action persists for a duration of $10$ seconds before the next action can be chosen.
To ensure safety, every transition from one phase to another involves a $3$-second yellow-change interval followed by $2$-second all-red interval to clear the intersection.

For our random training environments, we simulated passenger cars as vehicles only.
The same is true for \emph{cologne1}, \emph{cologne3} and \emph{cologne8}.
In addition to vanilla passenger cars, \emph{ingolstadt1}, \emph{ingolstadt7} and \emph{ingolstadt21} also include buses.

\subsubsection{Segment Length}
We choose a segment length $ds$ of $10$ meters for our experiments.
In theory, $ds$ is only upper bounded to $\ell$.
However, we argue in favour of tighter bounds in practice. 
Using segment densities as inputs factors out the number of vehicles and enables the encoder to learn segment lengths implicitly (if needed).
This requires a globally consistent segment length $ds$ (as mentioned in \cref{sec:segmentation}).
Hence, we drop possible remainders of $\frac{\ell}{ds}$.
The impact is minor for any reasonable choices of $ds$, as the cut-off is done at the end of the lane (maximally distant to the intersection). 
This approach introduces a technical upper bound to the choice of $ds$, as short lanes with $\ell < ds$ would be dropped.
Furthermore, choosing $ds$ smaller than the average vehicle length $h$ does not offer more valuable information for the encoding.
Therefore, we have $h \leq ds \leq \min_{\ell \in \mathcal{I} \cup \mathcal{O}} \ell$.

\begin{figure}
    \centering
    \includegraphics[width=1.\columnwidth]{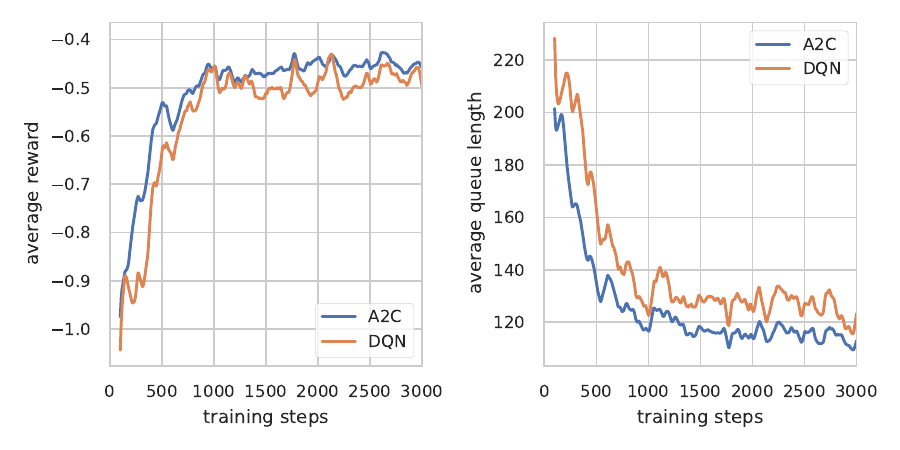}
    \caption{Average $\log$-pressure reward $(\uparrow)$ and the average queue length $(\downarrow)$ over $3000$ training steps.}
    \label{fig:training}
\end{figure}

\subsubsection{Training Details}
For optimisation, AdamW \cite{AdamW} with a learning rate of $1e-3$ and otherwise default settings is utilised.
Furthermore, we used mini-batches of $64$ SAR (state, action, reward) samples. 
We operate within a finite horizon of $0 \geq t \geq T$.
We also include a convergence illustration in \cref{fig:training}.
We found that both model versions converge within $3000$ steps (as the performance stays within reasonable error-bounds constant afterwards).
We skipped the first $100$ steps to let the traffic spawn in the simulation and develop a natural flow.

\subsubsection{Baseline Details}
All heuristics (incl. Random, FixedTime, and MaxPressure) are custom implementations.
All trainable baselines and related performance results are obtained using LibSignal \cite{LibSignal}.
Nonetheless, we used the same routines to compute the high-level performance indicators presented in our performance plots.

\subsection{Theory}

\subsubsection{Runtime Complexity}
Signals on the segment level are embedded in parallel and aggregated by \cref{eq:aggregation}.
Let $S_v := \sum_{\ell \in \mathcal{I}_v \cup \mathcal{O}_v} \frac{\ell}{ds}$ be the number of segments at $v$.
Due to the quadratic complexity of the attention mechanism in each of the three encoding layers, we get
\begin{equation} \label{eq:complexity}
    \mathcal{O} \left(S_v |\mathcal{M}_v| +|\mathcal{M}_v| |\Phi_v| + \Phi_v^2 + P_v \right),
\end{equation}
where $P_v := \sum_{\ell \in \mathcal{I}_v \cup \mathcal{O}_v} |\mathcal{M}_{\rightarrow \ell}| + |\mathcal{M}_{\leftarrow \ell}|$ is the number of movements required to estimate the transition priors in \cref{eq:prior}.
Note that $P_v \ll |\mathcal{M}|$ as only the prior only considers movements connecting two intersections.
For a reasonable segment length $ds$, $S_v |\mathcal{M}_v|$ is the dominant term in \cref{eq:complexity}.
As each intersection is controlled by weight-tied agents, policies can be computed in parallel for all $v$, disentangling the number of intersections from the runtime complexity.

\subsection{Ablation Studies}

\begin{figure*}
    \centering
    \includegraphics[width=1.\textwidth]{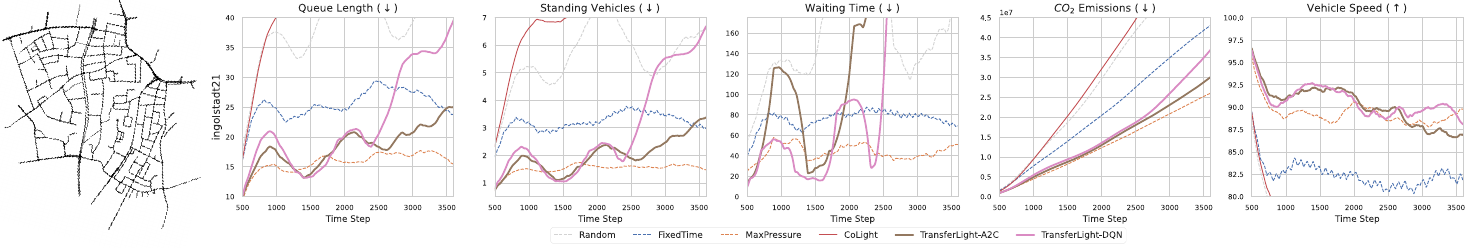}
    \caption{Test performances (moving averages) on \emph{Ingolstadt21} over 3600 simulated time steps.
    This is an example of the limits of generalisation capabilities of \emph{TransferLight}. Around step $2400$ a congestion is builds up around a few intersections which miscalculated some phase energies. Afterwards, it was not able to resolve the knot and the congestion spread across the network.
    }
    \label{fig:generalisation_fail}
\end{figure*}

\subsubsection{Reward Comparison}
\cref{fig:reward} compares the performance gains through our symmetry-breaking $\log$-distance reward.
We found that our $\log$-distance reward improves all three target performance indicators over the simulated test span.
These empirical results underpin our theoretical claims in \cref{sec:reward}.

\begin{figure}
\centering
\begin{tabular}{cccccc}
\multirow{6}{*}{\includegraphics[width=0.2\columnwidth]{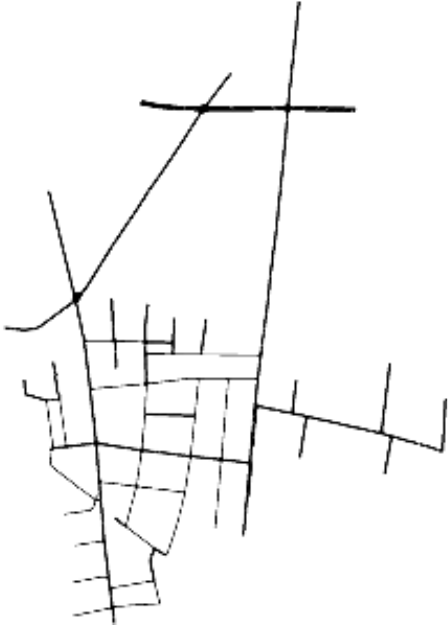}}
& $\operatorname{pe}$ & $\gamma_{m\phi}$ & $J_{\phi\phi^\prime}$ & $\bar{\rho}_\ell$ & ATT ($\downarrow$) \\
& & \checkmark & \checkmark & \checkmark & $92.13$ \small{$\pm 58.18$} \\
& \checkmark & & \checkmark & \checkmark & $144.52$ \small{$\pm 107.83$} \\
& \checkmark & \checkmark & & \checkmark & $92.15$ \small{$\pm 58.22$} \\
& \checkmark & \checkmark & \checkmark & & $92.80$ \small{$\pm 58.23$} \\
& \checkmark & \checkmark & \checkmark & \checkmark & $\mathbf{91.71}$ \small{$\pm \mathbf{56.51}$} \\
\end{tabular}
\caption{Ablation study of positional encoding $\operatorname{pe}$, edge features $\gamma_{m\phi}$ and $J_{\phi\phi^\prime}$, and our transition prior $\bar{\rho}_\ell$ comparing Average Travel Time (ATT) on \emph{cologne8}. Left: visualization of the cologne8 network. Right: ablation results.}
\label{tab:ablation_study}
\end{figure}

\subsubsection{Impact of Positional Encoding}
When lane segments are relatively short on average, the inclusion of positional encoding at the segment level has limited influence on model performance. This condition is particularly prevalent in urban environments, where signalized intersections are typically located in close proximity to one another. Under such circumstances, positional encoding can be effectively disregarded. Evidence supporting this claim is presented in \cref{tab:ablation_study}, where the omission of positional encoding resulted in only a negligible increase in average travel time.

\subsubsection{Impact of Transition Prior}
The removal of the state transition prior during both training and testing resulted in the second-largest performance degradation observed in our ablation study (see \cref{tab:ablation_study}). The transition prior plays a pivotal role in enabling the model to make proactive decisions rather than merely reactive ones, a capability that is especially critical in dense, urban environments. However, the observed impact of this prior was less substantial than initially anticipated. This discrepancy warrants further investigation, and we plan to explore the underlying reasons in greater detail in future work.

\subsubsection{Impact of Edge Features}
To introduce a measure of similarity between traffic signal phases (in terms of their green signal overlap), we employed the Jaccard index at the phase-to-phase level. As demonstrated in \cref{tab:ablation_study}, this feature provided only a marginal improvement in performance. At the movement-to-phase level, we incorporated edge feature encodings to represent whether a movement was prohibited, protected, or permitted. By a wide margin, this feature had the most significant impact on overall performance. We hypothesize that this added inductive bias enables the model to cluster movements more effectively before decoding them into phase energy distributions. Consequently, the model focuses primarily on traffic densities that can be directly alleviated by selecting specific phases.
Investigating this claim more thoroughly is another key point for potential follow-up work.

\begin{figure}
    \centering
    \includegraphics[width=1.\columnwidth]{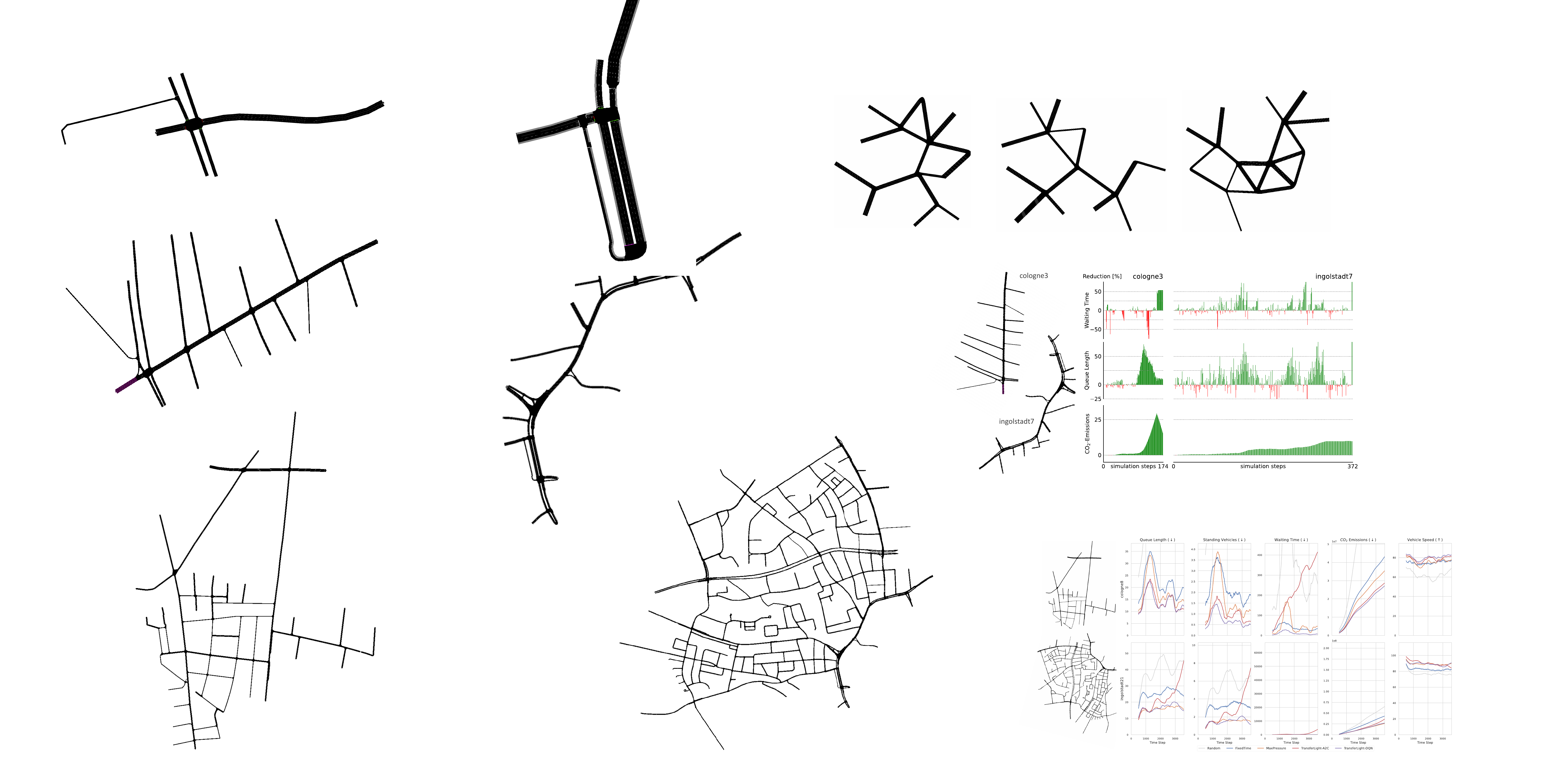}
    \caption{Waiting time, queue length, and emission reduction using our $\log$-distance pressure reward (\cref{eq:log_distance_reward}) compared to the commonly used pressure reward (\cref{eq:pressure}).
    }
    \label{fig:reward}
\end{figure}

\subsection{Limits of Generalisability}
The ability to generalise is of course facing limits at some range of problem complexity.
We performed an additional experiment on \emph{ingolstadt21} comprising $21$ intersections in a narrow urban environment.
\Cref{fig:generalisation_fail} compares our method to various baselines under different performance measures on this benchmark scenario.
After around 1200 time steps, \emph{TransferLight} with either head starts diverging into a suboptimal sequence of phases.
On the long run, this leads to congestions, which in turn lead to performance decreases among all measures.
We dedicate our future work to indicate the causing factors and prevent such situations to occur (under reasonable traffic demands).

\end{document}